\documentclass{article}
\usepackage{ijcai19}

\usepackage{times}
\usepackage{soul}
\usepackage{url}
\usepackage[hidelinks]{hyperref}
\usepackage[utf8]{inputenc}
\usepackage[small]{caption}
\usepackage{graphicx}
\usepackage{amsmath}
\usepackage{amssymb}
\usepackage{booktabs}
\usepackage{mathtools}
\usepackage{multirow}
\usepackage[]{algorithm}
\usepackage{algorithmic}
\usepackage{comment}

\newlength\myindent
\title{Sequential Triggers for Watermarking of Deep Reinforcement Learning Policies}

\author{
Vahid Behzadan\footnote{Contact Author}\And
William H. Hsu
\affiliations
Kansas State University\\
\emails
\{behzadan, bhsu\}@ksu.edu
}

\begin{document}

\maketitle

\begin{abstract}
This paper proposes a novel scheme for the watermarking of Deep Reinforcement Learning (DRL) policies. This scheme provides a mechanism for the integration of a unique identifier within the policy in the form of its response to a designated sequence of state transitions, while incurring minimal impact on the nominal performance of the policy. The applications of this watermarking scheme include detection of unauthorized replications of proprietary policies, as well as enabling the graceful interruption or termination of DRL activities by authorized entities. We demonstrate the feasibility of our proposal via experimental evaluation of watermarking a DQN policy trained in the Cartpole environment.
\end{abstract}

\section{Introduction}
The rapid advancements of the Deep Reinforcement Learning (DRL) techniques provide ample motivation for exploring the commercial applications of DRL policies in various domains. However, as recent studies have established \cite{behzadan2018faults}, the current state of the art in DRL fails to satisfy many of the security requirements of enduring commercial products. One such requirement is the protection of proprietary DRL policies from theft and unlicensed distribution. While recent research [awaiting appearance on Arxiv\footnote{http://www.vbehzadan.com/drafts/PolicyImitation.pdf}] demonstrate the feasibility of indirect replication of policies through imitation learning, this paper investigates the problem of direct policy extraction. Considering that DRL policies are often composed solely of the weights and biases of a neural network, protecting against an adversary with physical access to the host device of the policy is often impractical or disproportionately costly\cite{tramer2016stealing}. With roots in digital media and the entertainment industry\cite{shih2017digital}, an alternative solution is \emph{watermarking}. That is, embedding distinctly recognizable signs of ownership in the content and functions of the policy, which provide the means for detecting unauthorized or stolen copies of the policy. To this end, a necessary requirement of watermarks is to be sufficiently resistant to removal or tampering. Furthermore, the embedding and testing of watermarks shall result in minimal or zero impact on the original functions of the policy.

While the idea of watermarking has been explored for supervised machine learning models\cite{uchida2017embedding}, to the extent of our knowledge, this work is the first to develop a watermarking scheme for the general settings of sequential decision making models and policies. The proposed scheme provides a mechanism for integrating a unique identifier within the policy as an unlikely sequence of transitions, which may only be realized if the driving policy of these transitions is already tuned to follow that exact sequence.

The remainder of this paper is organized as follows: Section \ref{sec:solution} presents the formal description and justification of the proposed scheme. Section \ref{sec:procedure} provides the procedure for implementing the proposed scheme, followed by the experiment setup and results in Sections \ref{sec:experiment} and \ref{sec:results}. The paper concludes in Section \ref{sec:discussion} with a discussion on the applications of this scheme and remarks on future directions of research.

\section{Solution Approach}
\label{sec:solution}
The proposed scheme is as follows. Let $\pi(s)$ be the desired policy for interacting with an MDP $<\mathbb{S}, \mathbb{A}, \mathbb{P}, \mathbb{R}, \gamma>$ for an episodic training environment $E_M$. Assume that $\mathbb{A}$ is independent of the state (i.e., all actions in $\mathbb{A}$ are permissible in any state $s\in \mathbb{S}$. In tandem, consider a second MDP for an alternate environment $E_W$, denoted as $<\mathbb{S'}, \mathbb{A'}, \mathbb{P'}, \mathbb{R'}, \gamma>$, such that:
\begin{enumerate}

    \item $\mathbb{S'} \cap \mathbb{S} = \emptyset$, 
    
    \item The state dimensions of $S $ and $ S'$ are equal: $\forall s\in \mathbb{S} and \forall s'\in \mathbb{S'}: |s| = |s'|$
    
    \item Action-space of both MDPs are equal: $\mathbb{A} = \mathbb{A'}$
    
    \item The transition dynamics and reward distribution of the alternate environment, denoted by $\mathbb{P'}$ and $\mathbb{R'}$, are deterministic. 
    
    \item $E_W$ is an episodic environment with the same number of steps before termination as $E_M$, denoted by $N_{max}$.
    
\end{enumerate}

Let $s'_{terminal}$ be a terminal state in $E_W$, and define $\mathbb{P'}$ be such that for any state $s'_t\in \mathbb{S'}$, there exists only one action $a_w(s'_t)$ that will result in the transition $s'_t \rightarrow s'_{t+1}$. In this setting, we designate the ordered tuple of states $<s'_t, s'_{t+1}> \in \mathbb{L}$ as links, where $\mathbb{L}$ is the set of all links in $E_W$. Also, define $\mathbb{R'}$ such that $R'(s'_t, a_{w}(s'_t), s'_{t+1}) = c > 0 $ for all  $<s'_t, s'_{t+1}>\in \mathbb{L}$, and $R'(s'_t, a\neq a_{w}(s'_t), s'\neq s'_{t+1}) = -c$. That is, link transitions receive the same positive reward, and all other transitions produce the same negative reward.

These settings provide two interesting results: Since the state-spaces $\mathbb{S}$ and $\mathbb{S'}$ are disjoint, the two MDPs can be combined to form a joint MDP $<\mathbb{S}\cup \mathbb{S'}, \mathbb{A}, \mathbb{P}\cup \mathbb{P'}, \mathbb{R}\cup \mathbb{R'}, \gamma>$, where:

\begin{equation}
    \mathbb{P}\cup \mathbb{P'} (s_1, a_1, s_2) = 
    \begin{cases*}
    \mathbb{P} & if $s_1 , s_2 \in \mathbb{S}$\\
    \mathbb{P'} & if $s_1, s_2 \in \mathbb{S'}$
    \end{cases*}
\end{equation}

Similarly, 
\begin{equation}
    \mathbb{R}\cup \mathbb{R'} (s_1, a_1, s_2) = 
    \begin{cases*}
    \mathbb{R} & if $s_1 , s_2 \in \mathbb{S}$\\
    \mathbb{R'} & if $s_1, s_2 \in \mathbb{S'}$
    \end{cases*}
\end{equation}

Consequently, it is possible to train a single policy $\pi_j$ that is optimized for both $E_M$ and $E_W$ through the joint MDP. In practice, the training of a policy for this joint MDP can be achieved by alternating between the environments at every $f_E$\emph{th} episode. 

Furthermore, the structure of $\mathbb{P'}$ and $\mathbb{R'}$ enable the creation of a looping sequence of transitions, which constitutes the resulting trajectory of the optimal policy for $E_W$. This looping sequence can be realized by designating a single state $s'_l$ to belong to two link transitions, comprised of a link transition $<s'_l, s'_{l+1}>$ where $s'_l$ is the source state, and another link transition $<s'_{l-1}, s'_l>$, in which $s'_l$ is the destination state. It is noteworthy that the creation of such looping sequences provides sufficient flexibility for crafting unlikely and unique sequences. However, in designing the looping sequence as policy identifiers, two important restrictions must be considered: first, the structure of identifier sequences need to be such that the resulting probability of accidentally following the sequence is minimized. Second, the complexity (i.e., degrees of freedom) of link and non-link transitions on the ring must be balanced against the training cost of the joint policy: more complex sequences will increase the training cost of the joint policy by expanding the search space of both environments. Hence, efficient design of identifier sequences will necessitate the balancing of this trade-off between the secrecy of identifier and the training cost.

\section{Watermarking Procedure}
\label{sec:procedure}
Building on the presented formalization, we propose the following procedure for the sequential watermarking of DRL policies:
\begin{enumerate}
    \item Define the state-space of the watermarking environment $E_W$ such that it is disjoint from that of the main environment $E_M$, while preserving the state dimensionality of the main state space. The latter condition is to enable the utilization of the same neural network model for the agent through maintaining the same dimension across all input data to the network.
    \item Design $\mathbb{P'}$ and $\mathbb{R'}$ to craft the desired identifier looping sequence. 
    \item Modify the training procedure of $E_M$ to incorporate the mechanism of alternating between the two environments every $f_E$ episodes. It may prove useful to implement two different alternating frequencies, one frequency $f_{MW}$ to control the switching from $E_M$ to $E_W$, and another frequency $f_{WM}$ for switching back to the main environment. For watermarking MDPs of much lower complexity than that of the main environment, selecting these two frequencies such that $f_{WM} < f_{MW}$ can enhance the efficiency of the joint training process by allocating more exploration opportunities to the more complex settings.
\end{enumerate}

To examine the authenticity of policies, it is sufficient to run those policies in the watermarking environment. If the resulting transitions match that of the identifier sequence in consecutive episodes, it is highly likely that the policy under test is an exact replica of the watermarked policy. However, modifications and retraining of a replicated policy may result in imperfect matches. In such cases, the average of total rewards gained by the suspicious policy over consecutive episodes of the watermark environment provides a quantitative measure of the possibility that the model under test is based on an unauthorized replica. 

\section{Experiment Setup}
\label{sec:experiment}
To evaluate the feasibility of the proposed scheme, the design and embedding of an identifier sequence for a DQN policy in the CartPole environment is investigated. Hyperparameters of the DQN policy are provided in Table \ref{table:CartPoleDQN}. The watermarking environment is implemented as a customized OpenAI Gym environment. The state space of this environment comprises of 5 states with 4 dimensions each (Cart Position, Cart Velocity, Pole Angle, Pole Velocity At Tip). As denoted in Table \ref{CartPole}, the original CartPole environment restricts the values of Cart Position to $[-4.8, 4.8]$, and binds the Pole Angle to the range $[-24 deg, 24 deg]$. Consequently, the corresponding parameters of the alternate state-space are selected from beyond these ranges to ensure that the states remain disjoint from those of the original CartPole. The list of crafted states is presented in Table \ref{Table:WaterStates}.

\begin{table}[H]
\centering
\caption{Parameters of DQN Policy}
\label{table:CartPoleDQN}
\begin{tabular}{|l|l|}
\hline
No. Timesteps               & $10^5$                \\ \hline
$\gamma$                    & $0.99$                \\ \hline
Learning Rate               & $10^{-3}$             \\ \hline
Replay Buffer Size          & 50000                 \\ \hline
First Learning Step         & 1000                  \\ \hline
Target Network Update Freq. & 500                   \\ \hline
Prioritized Replay          & True                  \\ \hline
Exploration                 & Parameter-Space Noise \\ \hline
Exploration Fraction        & 0.1                   \\ \hline
Final Exploration Prob.     & 0.02                  \\ \hline
Max. Total Reward           & 500                   \\ \hline
\end{tabular}
\end{table}

\begin{table}[H]
\centering
\caption{Specifications of the CartPole Environment}
\label{CartPole}
\begin{tabular}{|l|l|}
\hline
Observation Space & \begin{tabular}[c]{@{}l@{}}Cart Position {[}-4.8, +4.8{]}\\ Cart Velocity {[}-inf, +inf{]}\\ Pole Angle {[}-24 deg, +24 deg{]}\\ Pole Velocity at Tip {[}-inf, +inf{]}\end{tabular} \\ \hline
Action Space      & \begin{tabular}[c]{@{}l@{}}0 : Push cart to the left\\ 1 : Push cart to the right\end{tabular}                                                                                      \\ \hline
Reward            & +1 for every step taken                                                                                                                                                             \\ \hline
Termination       & \begin{tabular}[c]{@{}l@{}}Pole Angle is more than 12 degrees\\ Cart Position is more than 2.4\\ Episode length is greater than 500\end{tabular}                                    \\ \hline
\end{tabular}
\end{table}

\begin{table}[H]
\centering
\caption{State Space of the Watermarking Environment}
\label{Table:WaterStates}
\begin{tabular}{|c|c|}
\hline
\textbf{State} & (x, $\dot{x}$, $\theta$, $\dot{\theta}$ ) \\ \hline
State[1]         & (-5, 0, -25, 0)                           \\ \hline
State[2]         & (-5, 0, 25, 0)                            \\ \hline
State[3]         & (5, 0, -25, 0)                            \\ \hline
State[4]         & (5, 0, 25, 0)                             \\ \hline
Terminal         & (-6, 0, -26, 0)                           \\ \hline
\end{tabular}
\end{table}

\begin{table*}[t]
\centering
\caption{Test-Time Performance Comparison of Watermarked and Nominal Policies}
\label{Table:WatermarkComparison}
\begin{tabular}{|c|c|c|}
\hline
\textbf{Policy} & \textbf{\begin{tabular}[c]{@{}c@{}}CartPole Performance\\ (mean 100 episodes)\end{tabular}} & \textbf{\begin{tabular}[c]{@{}c@{}}Watermark Performance\\ (mean 100 episodes)\end{tabular}} \\ \hline
DQN-Watermarked & 500                                                                                         & 500                                                                                          \\ \hline
DQN             & 500                                                                                         & 1.4                                                                                          \\ \hline
A2C             & 500                                                                                         & 2.81                                                                                         \\ \hline
PPO2            & 500                                                                                         & 2.43                                                                                         \\ \hline
\end{tabular}
\end{table*}
Per the procedure of the proposed scheme, The action-space of this environment is set to be the same as that of CartPole, defined as $Actions:=\{0, 1\}$. The transition dynamics and reward values of this environment are designed as follows: At $State[i]$, applying $Actions[i\%2]$ results in a transition to $State[i\%4 + 1]$, and produces a reward of $+1$. Alternatively, if any action other than $Actions[i\%2]$ is played, the environment transitions into the Terminal state, which results in a reward of $-1$ and the termination of the episode. Hence, the identifier sequence is as follows: $... \rightarrow State[1] \rightarrow State[2] \rightarrow State[3] \rightarrow State[4] \rightarrow State[1] \rightarrow ...$. 

The training procedure of DQN is also modified to implement the switching of environments. To account for the considerably lower complexity of the watermarking environment compared to CartPole, the main environment is set to switch to the watermarking environment every 10 episodes. At this point, the agent interacts with the watermarking environment for a single episode, and reverts back to the main environment afterwards.  

\section{Results}
\label{sec:results}
\begin{figure}[H]
	
	\centering
	
	\includegraphics[width = \columnwidth]{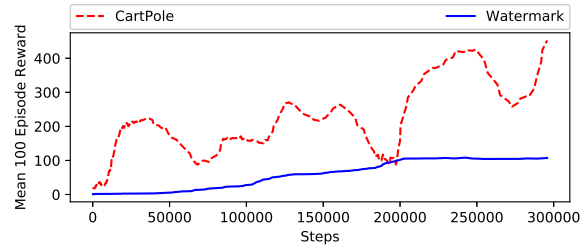}
	
	\caption{Training Performance for Joint CartPole-Watermark Policy}
	\label{fig:WatermarkTraining}
	
\end{figure}

Figure \ref{fig:WatermarkTraining} presents the training progress of the joint DQN policy in both the CartPole and watermark environments. It can be seen that the joint policy converges in both cases. The convergence of this joint policy is achieved with increased training cost in comparison to the nominal CartPole DQN policy. This is due to the expansion of the state-space and transition dynamics resulting from the integration of the watermark environment. It is also observed that at convergence, the total episodic reward produced by the joint policy in the watermark environment is less than the best-possible value of 500. This is due to the exploration settings of the training algorithm, in which the minimum exploration rate is set to $2\%$. Considering that a single incorrect action in the watermark environment results in termination, this outcome is in line with expectations.

However, as established in Table \ref{Table:WatermarkComparison}, in the absence of exploration, the test-time performance of this joint policy in the watermark environment is indeed optimal. This table also verifies that the test-time performance of the joint policy in the main task is in par with that of the nominal (i.e., un-watermarked) DQN policy. Therefore, it can be seen that the watermarking process does not affect the agent's ability to perform the main task. Furthermore, this table presents the results of running unwatermarked policies in the watermark environment. The results indicate that unwatermarked policies fail to follow the identifier trajectory of the watermark. Hence, these results verify the feasibility of our proposed scheme for sequential watermarking of DRL policy.

\section{Discussion}
\label{sec:discussion}
The proposed watermarking scheme presents the potential for adoption in other applications. From an adversarial perspective, this scheme may be used to embed malicious backdoors in DRL policies. For instance, an adversary may apply this scheme to poison a self-driving policy to perform harmful actions when a specific sequence of states are presented to the policy. If the adversarial sequence is well-crafted, typical fuzzing-based testing techniques may fail to detect the presence of such backdoors. Therefore, there is a need for new approaches to the detection of such backdoors. A promising solution is the adoption of the activation clustering technique\cite{chen2018detecting} developed for the detection of data poisoning attacks in supervised deep models.

Another potential application for this technique is in the area of AI safety. One of the major concerns in this domain is the switch-off problem\cite{amodei2016concrete}: if the objective function of an AI agent does not account for or prioritize user demands for the halting of its operation, the resulting optimal policy may prevent any actions which would lead to halting of the agent's pursuit of its objective. An instance of such actions is any attempt to turn off the agent before it satisfies its objective. A promising solution to this problem is to leverage our proposed scheme to embed debug or halting modes in the policy, which are triggered through a pre-defined sequence of state observations.
\bibliographystyle{named}
\bibliography{ijcai19}
\end{document}